\documentclass[letterpaper, 10pt, conference]{ieeeconf} 

\pdfoutput=1

\IEEEoverridecommandlockouts                              

\usepackage{times}

\usepackage{multicol}
\usepackage[bookmarks=true]{hyperref}
\usepackage{amsfonts}
\usepackage{mathrsfs}
\usepackage{amsmath}
\usepackage{bm}
\usepackage{yfonts}
\usepackage{amssymb}
\usepackage{breqn}
\usepackage{algorithm}
\usepackage{algpseudocode}
\usepackage{wrapfig}
\usepackage{subcaption}
\usepackage{graphics} 
\usepackage{epsfig} 
\usepackage{multirow}
\usepackage{verbatim}

\usepackage{booktabs,color}

\newcommand\at[2]{\left.#1\right|_{#2}}

\newcommand{\citet}[1]{\cite{#1}}

\title{\LARGE \bf  Direct Sparse Visual-Inertial Odometry using Dynamic Marginalization}

\author{Lukas von Stumberg$^{1}$, Vladyslav Usenko$^{1}$, Daniel Cremers$^{1}$%
\thanks{$^{1}$The authors are with the Computer Vision Group, Computer Science Institute 9, Technische Universit\"at M\"unchen, 85748 Garching, Germany
        {\tt\small \{stumberg, usenko, cremers\}@in.tum.de}}%
}

\begin{document}

\thispagestyle{empty} 
\onecolumn 
 
\begin{center} 
\noindent 
 
This paper has been accepted for publication in \emph{2018 International Conference on Robotics and Automation}. 
 
\end{center} 
\vspace{3em} 
 
\copyright2018 IEEE. Personal use of this material is permitted. Permission from IEEE must be obtained for all other uses, in any current or future media, including reprinting/republishing this material for advertising or promotional purposes, creating new collective works, for resale or redistribution to servers or lists, or reuse of any copyrighted component of this work in other works.
 
\twocolumn

\setcounter{page}{1}

\maketitle
\thispagestyle{empty}
\pagestyle{empty}

\begin{abstract}
We present VI-DSO, a novel approach for visual-inertial odometry,
which jointly estimates camera poses and sparse scene
geometry by minimizing photometric and IMU measurement
errors in a combined energy functional. The visual part of
the system performs a bundle-adjustment like optimization
on a sparse set of points, but unlike key-point based
systems it directly minimizes a photometric error. This
makes it possible for the system to track not only corners,
but any pixels with large enough intensity gradients. IMU
information is accumulated between several frames using
measurement preintegration, and is inserted into the
optimization as an additional constraint between keyframes.
We explicitly include scale and gravity direction into our
model and jointly optimize them together with other
variables such as poses. As the scale is often not
immediately observable using IMU data this allows us to
initialize our visual-inertial system with an arbitrary
scale instead of having to delay the initialization until
everything is observable. We perform partial
marginalization of old variables so that updates can be
computed in a reasonable time. In order to keep the system
consistent we propose a novel strategy which we call
"dynamic marginalization". This technique allows us to use
partial marginalization even in cases where the initial
scale estimate is far from the optimum. We evaluate our
method on the challenging EuRoC dataset, showing that VI-DSO outperforms the state of the art.
\end{abstract}

\section{Introduction}

\begin{figure}[htb]
    \centering
    \includegraphics[width=\linewidth]{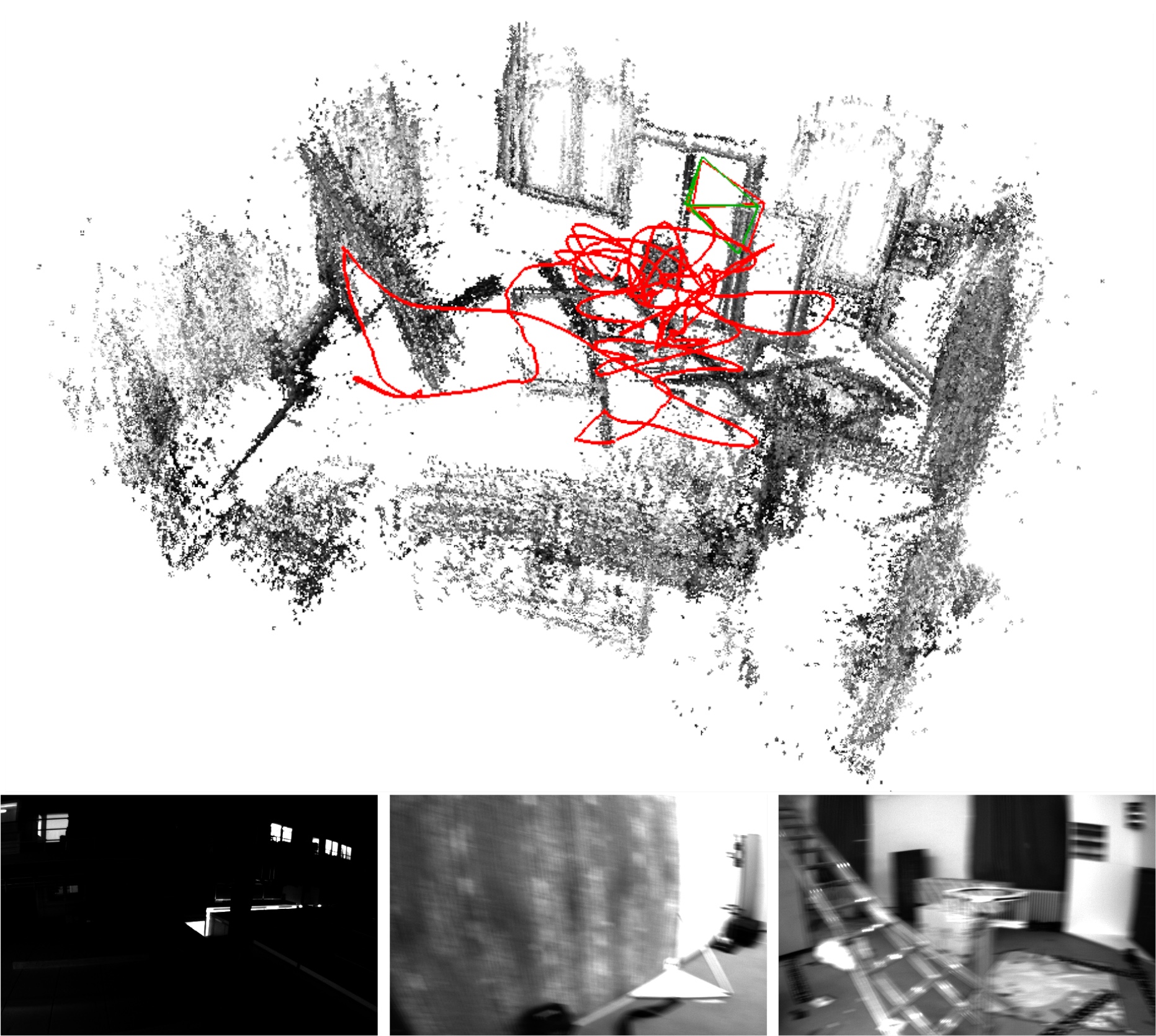}
    \caption{Bottom: Example images from the EuRoC-dataset: Low illumination, strong motion blur and little texture impose significant challenges for odometry estimation. Still our method is able to process all sequences with a rmse of less then 0.23m. Top: Reconstruction, estimated pose (red camera) and groundtruth pose (green camera) at the end of V1\_03\_difficult.}
    \label{fig:teaser}
\end{figure}

Motion estimation and 3D reconstruction are crucial tasks for robots. In general, many different sensors can be used for these tasks: laser rangefinders, 
RGB-D cameras \cite{kerl13icra}, GPS and others. 
Since cameras are cheap, lightweight and small passive sensors they have drawn a large attention of the community. Some examples of practical applications include robot navigation
\cite{stelzer12ijrr} and (semi)-autonomous driving \cite{geiger12cvpr}. 
However, current visual odometry methods suffer from a lack of robustness when confronted with low textured areas or fast maneuvers.
To eliminate these effects a combination with another passive sensor - an inertial measurement unit ~(IMU) can be used. It provides accurate short-term motion constraints and, unlike vision, is not prone to outliers. 

In this paper we propose a tightly coupled direct approach to visual-inertial odometry. It is based on Direct Sparse Odometry ~(DSO) \cite{engel2016dso} and uses a bundle-adjustment like photometric error function that simultaneously optimizes 3D geometry 
and camera poses in a combined energy functional. We complement the error function with IMU measurements. This is particularly beneficial for direct methods, since the error function is highly non-convex and a good initialization is important. 
A key drawback of monocular visual odometry is that it is not possible to obtain the metric scale of the environment. Adding an IMU enables us to observe the scale. Yet, depending on the performed motions this can take infinitely long, making the initialization a challenging task. 
Rather than relying on a separate IMU initialization we include the scale as a variable into the model of our system and jointly optimize it together with the other parameters.

Quantitative evaluation on the EuRoC dataset \cite{Burri2016Euroc} demonstrates that we can reliably determine camera motion and sparse 3D structure (in metric units) from a visual-inertial system on a rapidly moving micro aerial vehicle (MAV) despite challenging illumination conditions (Fig. \ref{fig:teaser}).

In summary, our contributions are:
\begin{itemize}
    \item a direct sparse visual-inertial odometry system. 
    \item a novel initialization strategy where scale and gravity direction are included into the model and jointly optimized after initialization.
    \item we introduce "dynamic marginalization" as a technique to adaptively employ marginalization strategies even in cases where certain variables undergo drastic changes.
    \item an extensive evaluation on the challenging EuRoC dataset showing that both, the overall system and the initialization strategy outperform the state of the art.
\end{itemize}

\section{Related work}

Motion estimation using cameras and IMUs has been a popular research topic for many years. In this section we will give a summary of visual, and visual-inertial odometry methods. 
We will also discuss approaches to the initialization of monocular visual-inertial odometry, where the initial orientation, velocity and scale are not known in advance.

The term visual odometry was introduced in the work of Nister et al. \cite{nister2004vo}, who proposed to use frame-to-frame matching of the sparse set of points to estimate the motion of the cameras. Most of the early approaches were based on matching features detected in the images, in particular MonoSLAM \cite{davison07pami}, a real-time capable EKF-based method. Another prominent example is PTAM \cite{klein07ismar}, which combines a bundle-adjustment backend for mapping with real-time capable tracking of the camera relative to the constructed map. Recently, a feature-based system capable of large-scale real-time SLAM was presented by Mur-Artal et al. \cite{murartal15orbslam}.

Unlike feature-based methods, direct methods use un-processed intensities in the image to estimate the motion of the camera. The first real-time capable direct approach for stereo cameras was presented in \cite{comport07icra}. 
Several methods for motion estimation for RGB-D cameras were developed by Kerl et al. \cite{kerl13icra}. 
More recently, direct approaches were also applied to monocular cameras, in a dense \cite{newcombe2011iccv}, semi-dense \cite{engel14eccv}, and sparse fashion \cite{Forster14} \cite{engel2016dso}.

Due to the complementary nature of the IMU sensors, there were many attempts to combine them with vision. They provide good short-term motion prediction and make roll and pitch angles observable. 
At first, vision systems were used just as a provider of 6D pose measurements which were then inserted in the combined optimization. 
This, so-called \emph{loosely coupled} approach, was presented in 
\cite{meier11icra} and \cite{engel12iros}. It is generally easier to implement, since the vision algorithm requires no modifications. On the other hand, \emph{tightly coupled} approaches jointly optimize motion parameters in a combined energy function. They are able to capture more correlations in the multisensory data stream leading to more precision and robustness. Several prominent examples are filtering based approaches \cite{mourikis13IJRR} \cite{Bloesch} 
and energy-minimization based approaches \cite{leutenegger2014keyframe} \cite{Forster-RSS-15} \cite{usenko16icra} \cite{murartal16vio}.

Another issue relevant for the practical use of monocular visual-inertial odometry is initialization. Right after the start, the system has no prior information about the initial pose, velocities and depth values of observed points in the image. Since the energy functional that is being minimized is highly non-convex, a bad initialization might result in divergence of the system. The problem is even more complicated, since some types of motion 
do not allow to uniquely determine all these values. 
A closed form solution for initialization, together with analysis of the exceptional cases was presented in \cite{Martinelli2014}, and extended to consider IMU biases in \cite{Kaiser14imu_init}.

\section{Direct Sparse Visual-Inertial Odometry}

The following approach is based on iterative minimization of photometric and inertial errors in a non-linear optimization framework. To make the problem computationally feasible the optimization is performed on a window of recent frames while all older frames get marginalized out. Our approach is based on \cite{engel2016dso} and can be viewed as a direct formulation of \cite{leutenegger2014keyframe}. In contrast to \cite{usenko16icra}, we jointly determine poses and 3D geometry from a single optimization function. This results in better precision especially on hard sequences. Compared to \cite{Forster-RSS-15} we perform a full bundle-adjustment like optimization instead of including structure-less vision error terms.

The proposed approach estimates poses and depths by minimizing the energy function
\begin{equation}
    E_{\text{total}} = \lambda \cdot E_{\text{photo}} + E_{\text{inertial}}
\end{equation}
which consists of the photometric error $E_{\text{photo}}$ (section \ref{sec:photometricerror}) and an inertial error term $E_{\text{inertial}}$ (section \ref{sec:inertialerror}).

The system contains two main parts running in parallel:
\begin{itemize}
    \item The coarse tracking is executed for every frame and uses direct image alignment combined with an inertial error term to estimate 
    the pose of the most recent frame.
    \item When a new keyframe is created we perform a visual-inertial bundle adjustment like optimization that estimates the geometry and poses of all active keyframes.
\end{itemize}
In contrast to \citet{murartal16vio} we do not wait for a fixed amount of time before initializing the visual-inertial system but instead we jointly optimize all parameters including the scale. This yields a higher robustness as inertial measurements are used right from the beginning.

\subsection{Notation}

Throughout the paper we will use the following notation: bold upper case letters $\mathbf{H}$ represent matrices, bold lower case $\bm{x}$ vectors and light lower case $\lambda$ represent scalars. Transformations between coordinate frames are denoted as $\mathbf{T}_{i\_j} \in \mathbf{SE}(3)$ where point in coordinate frame $i$ can be transformed to the coordinate frame $j$ using the following equation 
$\mathbf{p}_i = \mathbf{T}_{i\_j} \mathbf{p}_j$. We denote Lie algebra elements as $\hat{\bm{\xi}} \in \mathfrak{se}(3)$, where $\bm{\xi} \in \mathbb{R}^6$, and use them to apply small increments to the 6D pose $\bm{\xi}_{i\_j}' = \bm{\xi}_{i\_j} \boxplus \bm{\xi} :=  \log \left( e^{\hat{\bm{\xi}}_{i\_j}} \cdot e^{\hat{\bm{\xi}}} \right)^{\vee}$. 

We define the \emph{world} as a fixed inertial coordinate frame with gravity acting in negative $Z$ axis. 
We also assume that the transformation from camera to IMU frame $T_{\text{imu}\_\text{cam}}$ is fixed and calibrated in advance. 
Factor graphs are expressed as a set $G$ of factors and we use $G_1 \cup G_2$ to denote a factor graph containing all factors that are either in $G_1$ or in $G_2$.

\subsection{Photometric Error}\label{sec:photometricerror}

The photometric error of a point $p \in \Omega_i$ in reference frame $i$ observed in another frame $j$ is defined as follows:
\begin{align}\label{eq:photometric}
E_{\bm{p}j} &= \sum_{\mathbf{p} \in \mathcal{N}_{\bm{p}}} \omega_{\bm{p}} \bigg \lVert (I_j[\bm{p'}] - b_j) - \frac{t_j e^{a_j}}{t_i e^{a_i}} (I_i[\bm{p}] - b_i) \bigg \rVert_{\gamma},
\end{align}
where $\mathcal{N}_{\bm{p}}$ is a small set of pixels around the point $\bm{p}$, $I_i$ and $I_j$ are images of respective frames,  $t_i, t_j$ are the exposure times, $a_i, b_i, a_j, b_j$ are the coefficients to correct for affine illumination changes, $\gamma$ is the Huber norm, $\omega_p$ is a gradient-dependent weighting and $\bm{p'}$ is the point projected into $I_j$. 

With that we can formulate the photometric error as
\begin{align}
E_{\text{photo}} = \sum_{i \in \mathcal{F}} \sum_{\bm{p} \in \mathcal{P}_i} \sum_{j \in \text{obs}(\bm{p})} E_{\bm{p}j},
\end{align}
where $\mathcal{F}$ is a set of keyframes that we are optimizing, $\mathcal{P}_i$ is a sparse set of points in keyframe $i$, and $obs(\mathbf{p})$ is a set of observations of the same point in other keyframes.

\subsection{Inertial Error}
\label{sec:inertialerror}

In order to construct the error term that depends on rotational velocities measured by the gyroscope and linear acceleration measured by the accelerometer we use the nonlinear dynamic model defined in \cite[eq. (6), (7), (8)]{usenko16icra}.

    As IMU data is obtained with a much higher frequency than images we follow the preintegration approach proposed in \cite{Lupton12} and improved in \cite{carloneicra}  and \cite{Forster-RSS-15}. 
    This allows us to add a single IMU factor describing the pose between two camera frames.
    For two states $\bm{s}_{i}$ and $\bm{s}_{j}$ (based on the state definition in Equation (\ref{eq:imustate})), and IMU-measurements $\bm{a}_{i,j}$ and $\bm{\omega}_{i,j}$ between the two images we obtain a prediction $\widehat{\bm{s}}_j$ as well as an associated covariance matrix $\widehat{\Sigma}_{s,j}$.
    The corresponding error function is
    \begin{equation}
    E_{\text{inertial}}( \boldsymbol s_{i}, \boldsymbol s_j ) := \left(\boldsymbol s_j \boxminus \widehat{\boldsymbol s}_j \right)^T \mathbf{\widehat{\Sigma}}_{s,j}^{-1} \left(\boldsymbol s_j \boxminus \widehat{\boldsymbol s}_j \right)
    \end{equation}
    where the operator $\boxminus$ applies $\bm{\xi}_j \boxplus \left(\widehat{\bm{\xi}_j}\right)^{-1}$ for poses and a normal subtraction for other components.

\subsection{IMU Initialization and the problem of observability}

In contrast to a purely monocular system the usage of inertial data enables us to observe metric scale and gravity direction. This also implies that those values have to be properly initialized, otherwise optimization might diverge. Initialization of the monocular visual-inertial system is a well studied problem with an excellent summary provided in \cite{Martinelli2014}. \cite[Tables I and II]{Martinelli2014} show that for certain motions immediate initialization is not possible, for example when moving with zero acceleration and constant non-zero velocity. To demonstrate that it is a real-world problem and not just a theoretical case we note that the state-of-the-art visual-inertial SLAM system \cite{murartal16vio} uses the first 15 seconds of camera motion for the initialization on the EuRoC dataset to make sure that all values are observable.

Therefore we propose a novel strategy for handling this issue. We explicitly include scale (and gravity direction) as a parameter in our visual-inertial system and jointly optimize them together with the other values such as poses and geometry. This means that we can initialize with an arbitrary scale instead of waiting until it is observable.  
We initialize the various parameters as follows. 
\begin{itemize}
    \item We use the same visual initializer as \cite{engel2016dso} which computes a rough pose estimate between two 
    frames as well as approximate depths for several points. They are normalized so that the average depth is $1$.
    \item The initial gravity direction is computed by averaging up to $40$ accelerometer measurements, yielding a sufficiently good estimate even in cases of high acceleration.
    \item We initialize the velocity and IMU-biases with zero and the scale with $1.0$.
\end{itemize}
All these parameters are then jointly optimized during a bundle adjustment like optimization.

\subsection{$\mathbf{SIM}(3)$-based Representation of the World} \label{sec:sim3model} 
In order to be able to start tracking and mapping with a preliminary scale and gravity direction we need to include them into our model. Therefore in addition to the metric coordinate frame we define the DSO coordinate frame to be a scaled and rotated version of it. The transformation from the DSO frame to the metric frame is defined as $ \mathbf{T}_{m\_d} \in \{\mathbf{T} \in \mathbf{SIM}(3) \mid \text{translation}(\mathbf{T}) = 0 \}$, together with the corresponding $\bm{\xi}_{m\_d} = \log(\mathbf{T}_{m\_d}) \in \mathfrak{sim}(3)$. We add a superscript $D$ or $M$ to all poses denoting in which coordinate frame they are expressed.
In the optimization the photometric error is always evaluated in the DSO frame, making it independent of the scale and gravity direction, whereas the inertial error has to use the metric frame. 

\subsection{Scale-aware Visual-inertial Optimization} \label{sec:jointoptimization}

\begin{figure*}[t!]
    \centering
 \begin{subfigure}[t]{0.39\linewidth}
 \centering
        \includegraphics[height=3.7cm]{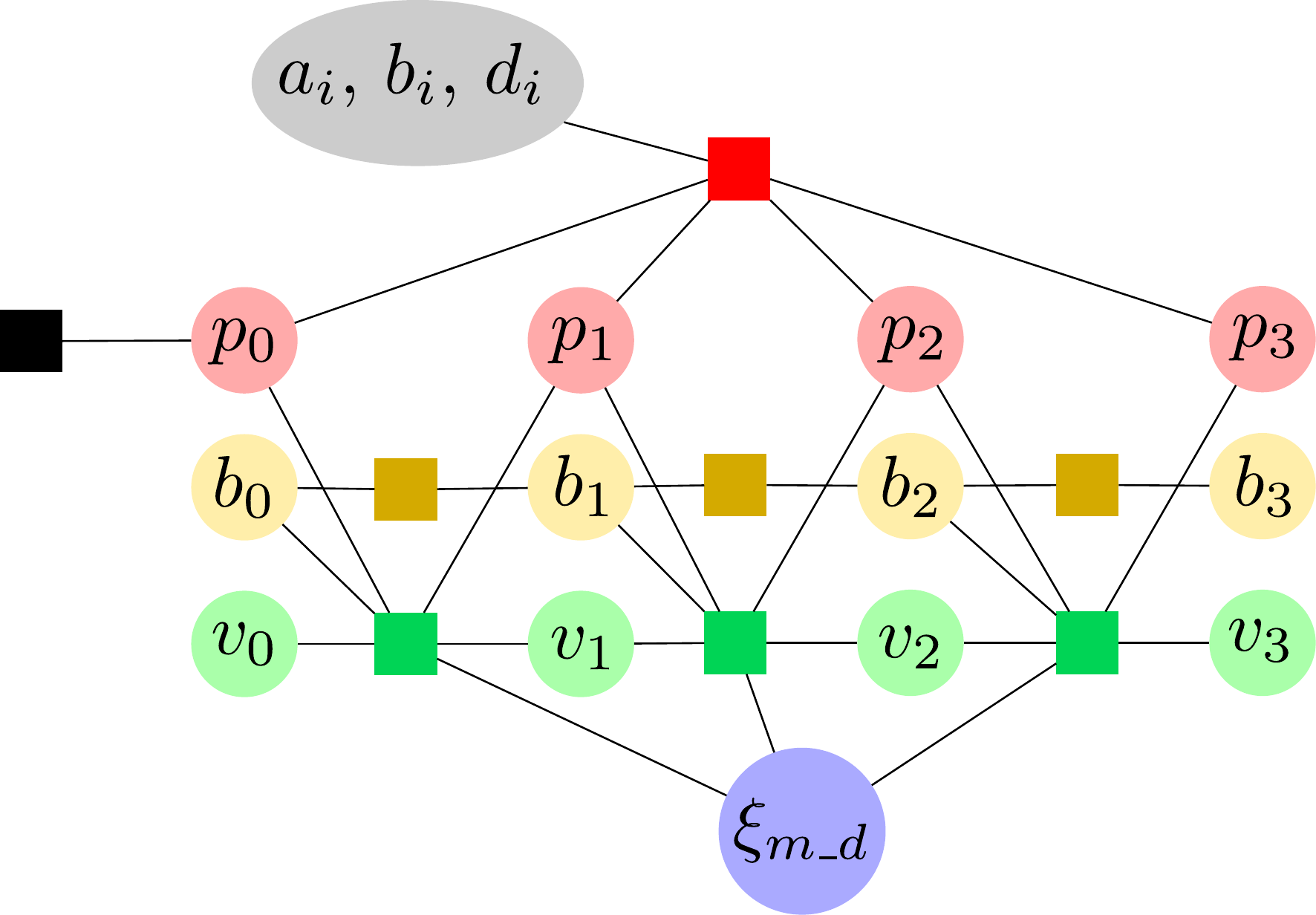} 
        \caption{Factor graph for the visual-inertial optimization.}
        \label{fig:bagraph}
    \end{subfigure}~
    \begin{subfigure}[t]{0.39\linewidth}
    \centering
        \includegraphics[height=3.7cm]{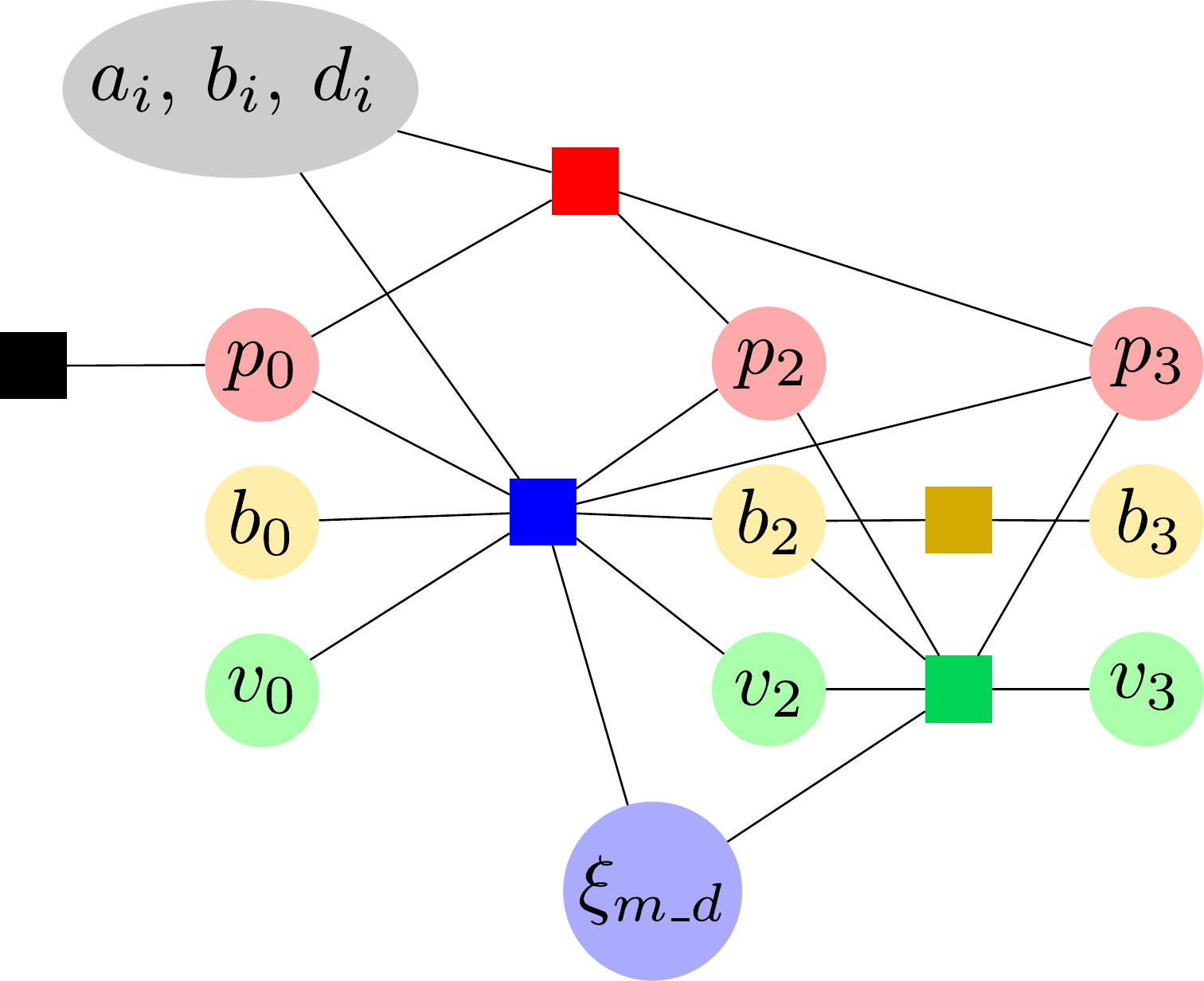}
        \caption{Factor graph after keyframe 1 was marginalized.}
        \label{fig:bagraphmarginalized}
    \end{subfigure}
    \begin{subfigure}[t]{0.2\linewidth}
        \centering
        \includegraphics[height=3.5cm]{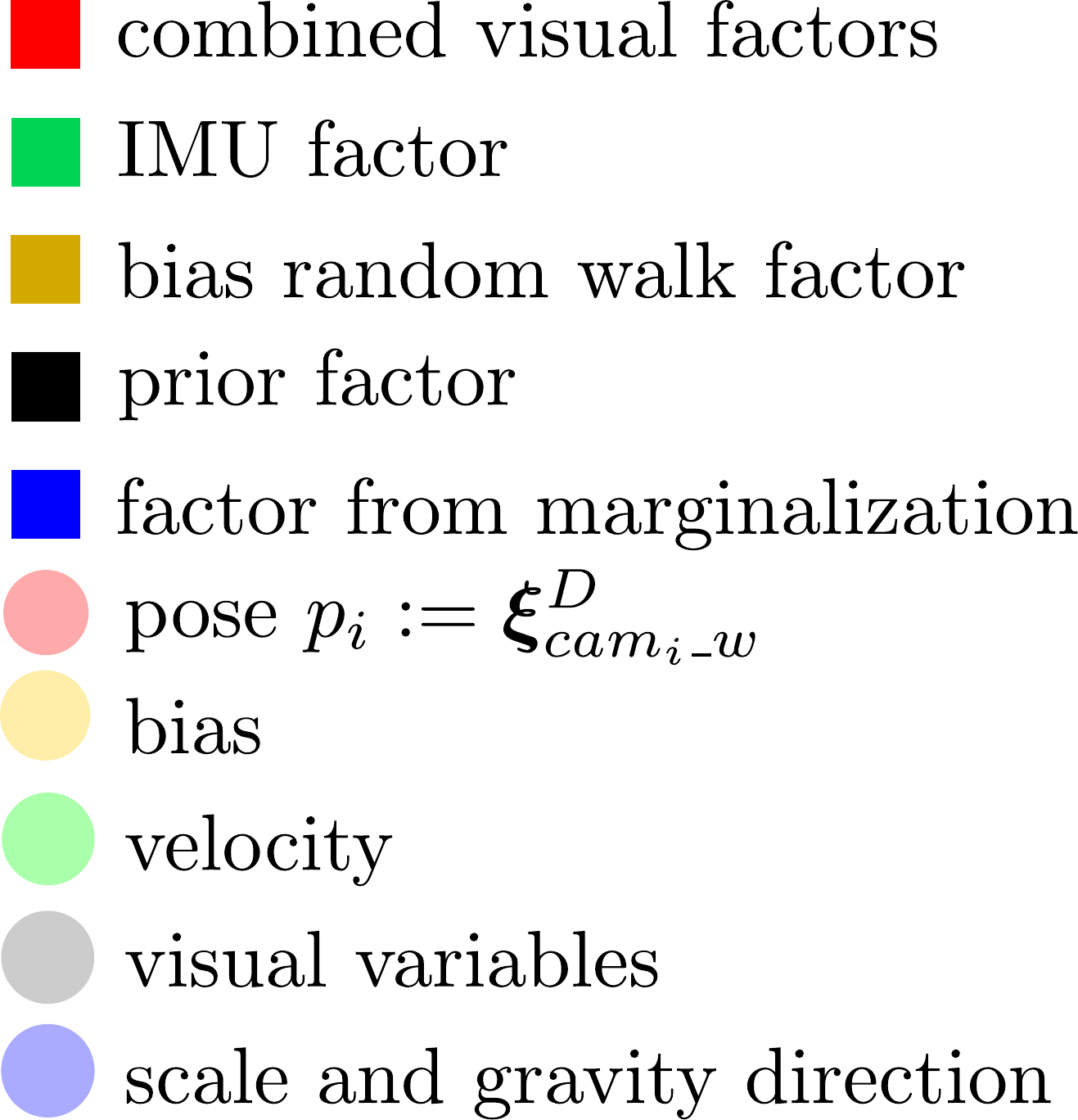}
        \label{fig:bagraphkey}
    \end{subfigure}
    \caption{Factor graphs for the visual-inertial joint optimization before and after the marginalization of a keyframe.}
	\label{fig:bagraphs}
\end{figure*}

We optimize the poses, IMU-biases and velocities of a fixed number of keyframes. Fig. \ref{fig:bagraph} shows a factor graph of the problem. Note that there are in fact many separate visual factors connecting two keyframes each, which we have combined to one big factor connecting all the keyframes in this visualization.
Each IMU-factor connects two subsequent keyframes using the preintegration scheme described in section \ref{sec:inertialerror}. As the error of the preintegration increases with the time between the keyframes we ensure that the time between two consecutive keyframes is not bigger than $0.5$ seconds which is similar to what \citet{murartal16vio} have done. 
Note that in contrast to their method however we allow the marginalization procedure described in section \ref{sec:marginalization} to violate this constraint which ensures that long-term relationships between keyframes can be properly observed.

An important property of our algorithm is that the optimized poses are not represented in the metric frame but in the DSO frame. This means that they do not depend on the scale of the environment. 

\subsubsection{Nonlinear Optimization}
We perform nonlinear optimization using the Gauss-Newton algorithm.
For each active keyframe we define a state vector 
\begin{equation}\bm{s}_i := [\left(\bm{\xi}_{cam_i\_w}^D\right)^T, \bm{v}_i^T, \bm{b}_i^T, a_i, b_i, d_i^1, d_i^2, ..., d_i^m]^T\end{equation}
 where $\bm{v}_i \in \mathbb{R}^3$ is the velocity, $\bm{b}_i \in \mathbb{R}^6$ is the current IMU bias, $a_i$ and $b_i$ are the affine illumination parameters used in equation (\ref{eq:photometric}) and $d_i^j$ are the inverse depths of the points hosted in this keyframe.
 
 The full state vector is then defined as
 \begin{equation} \label{eq:fullstate}
  \bm{s} = [\bm{c}^T, \bm{\xi}_{m\_d}^T, \bm{s}_1^T, \bm{s}_2^T, ..., \bm{s}_n^T]^T
 \end{equation}
 where $\bm{c}$ contains the geometric camera parameters and $\bm{\xi}_{m\_d}$ denotes the translation-free transformation between the DSO frame and the metric frame as defined in section \ref{sec:sim3model}. 
 We define the operator $\bm{s} \boxplus \bm{s}'$ to work on state vectors by applying the concatenation operation $\bm{\xi} \boxplus \bm{\xi'}$ for Lie algebra components and a plain addition for other components.

Using the stacked residual vector $\bm{r}$ we define 
\begin{equation}
 \mathbf{J} = \at{\frac{d \bm{r}\left( \bm{s} \boxplus \bm{\epsilon}\right)}{d \bm{\epsilon}}}{\bm{\epsilon}=0} \text{~,~} \mathbf{H}=\mathbf{J}^T \mathbf{W} \mathbf{J} \text{~and~} \bm{b}= -\mathbf{J}^T\mathbf{W}\bm{r}
\end{equation}

where $\mathbf{W}$ is a diagonal weight matrix.
Then the update that we compute is $\bm{\delta} = \mathbf{H}^{-1} \bm{b}$.

Note that the visual energy term $E_{\text{photo}}$ and the inertial error term $E_{\text{imu}}$ do not have common residuals. Therefore we can divide $\mathbf{H}$ and $\bm{b}$ each into two independent parts
\begin{equation}
 \mathbf{H} = \mathbf{H_{\text{photo}}} + \mathbf{H_{\text{imu}}} \text{~and~} \bm{b} = \bm{b_{\text{photo}}} + \bm{b_{\text{imu}}}
\end{equation}
 
 As the inertial residuals compare the current relative pose to the estimate from the inertial data they need to use poses in the metric frame relative to the IMU. Therefore we define additional state vectors 
 for the inertial residuals. 
 \begin{equation} \label{eq:imustate}
 \bm{s}_i' := [\bm{\xi}_{w\_imu_i}^M, \bm{v}_i, \bm{b}_i]^T \text{~and~}
  \bm{s}' = \left[\bm{s}_1^{\prime T}, \bm{s}_2^{\prime T}, ..., \bm{s}_n^{\prime T}\right]^T
 \end{equation}

The inertial residuals lead to 
\begin{equation}
    \mathbf{H}_{\text{imu}}' = \mathbf{J}_{\text{imu}}^{\prime T} \mathbf{W}_{\text{imu}} \mathbf{J}_{\text{imu}}'  \text{~and~}
     \bm{b}_{\text{imu}}' = -\mathbf{J}_{\text{imu}}^{\prime T}\mathbf{W}_{\text{imu}}\bm{r}_{\text{imu}}
\end{equation}
For the joint optimization however we need to obtain $\mathbf{H}_{\text{imu}}$ and $\bm{b}_{\text{imu}}$ based on the state definition in Equation (\ref{eq:fullstate}). As the two definitions mainly differ in their representation of the poses we can compute $\mathbf{J}_{\text{rel}}$ such that
\begin{equation}
    \mathbf{H}_{\text{imu}} = \mathbf{J}_{\text{rel}}^T \cdot \mathbf{H}_{\text{imu}}' \cdot \mathbf{J}_{\text{rel}} ~\text{and}~ \bm{b}_{\text{imu}} = \mathbf{J}_{\text{rel}}^T \cdot \bm{b}_{\text{imu}}'
\end{equation}
The computation of $\mathbf{J}_{\text{rel}}$ is detailed in the supplementary material.
Note that we represent all transformations as elements of $\mathfrak{sim}(3)$ and fix the scale to $1$ for all of them except $\bm{\xi}_{m\_d}$.

\subsubsection{Marginalization using the Schur-Complement} \label{sec:marginalization}
In order to compute Gauss-Newton updates in a reasonable time-frame we perform partial marginalization for older keyframes. This means that all variables corresponding to this keyframe (pose, bias, velocity and affine illumination parameters) are marginalized out using the Schur complement. 
Fig. \ref{fig:bagraphmarginalized} shows how marginalization changes the factor graph.\looseness=-1

The marginalization of the visual factors is handled as in \cite{engel2016dso} by dropping residual terms that affect the sparsity of the system and by first marginalizing all points in the keyframe before marginalizing the keyframe itself.

Marginalization is performed using the Schur-complement \cite[eq. (16), (17) and (18)]{engel2016dso}. As the factor resulting from marginalization requires the linearization point of all connected variables to remain fixed we apply \cite[eq. (15)]{engel2016dso} to approximate the energy around further linearization points.

In order to maintain consistency of the system it is important that Jacobians are all evaluated at the same value for variables that are connected to a marginalization factor as otherwise the nullspaces get eliminated. Therefore we apply "First Estimates Jacobians". For the visual factors we follow \cite{engel2016dso} and evaluate $\mathbf{J}_{\text{photo}}$ and $\mathbf{J}_{\text{geo}}$ at the linearization point.
When computing the inertial factors we fix the evaluation point of $\mathbf{J}_{\text{rel}}$ for all variables which are connected to a marginalization factor. Note that this always includes $\bm{\xi}_{m\_d}$.

\subsubsection{Dynamic Marginalization for Delayed Scale Convergence}\label{sec:dynamicmarginalization}

\begin{figure}[tb]
    \centering
 \begin{subfigure}[t]{0.47\linewidth}
        \includegraphics[width=\linewidth]{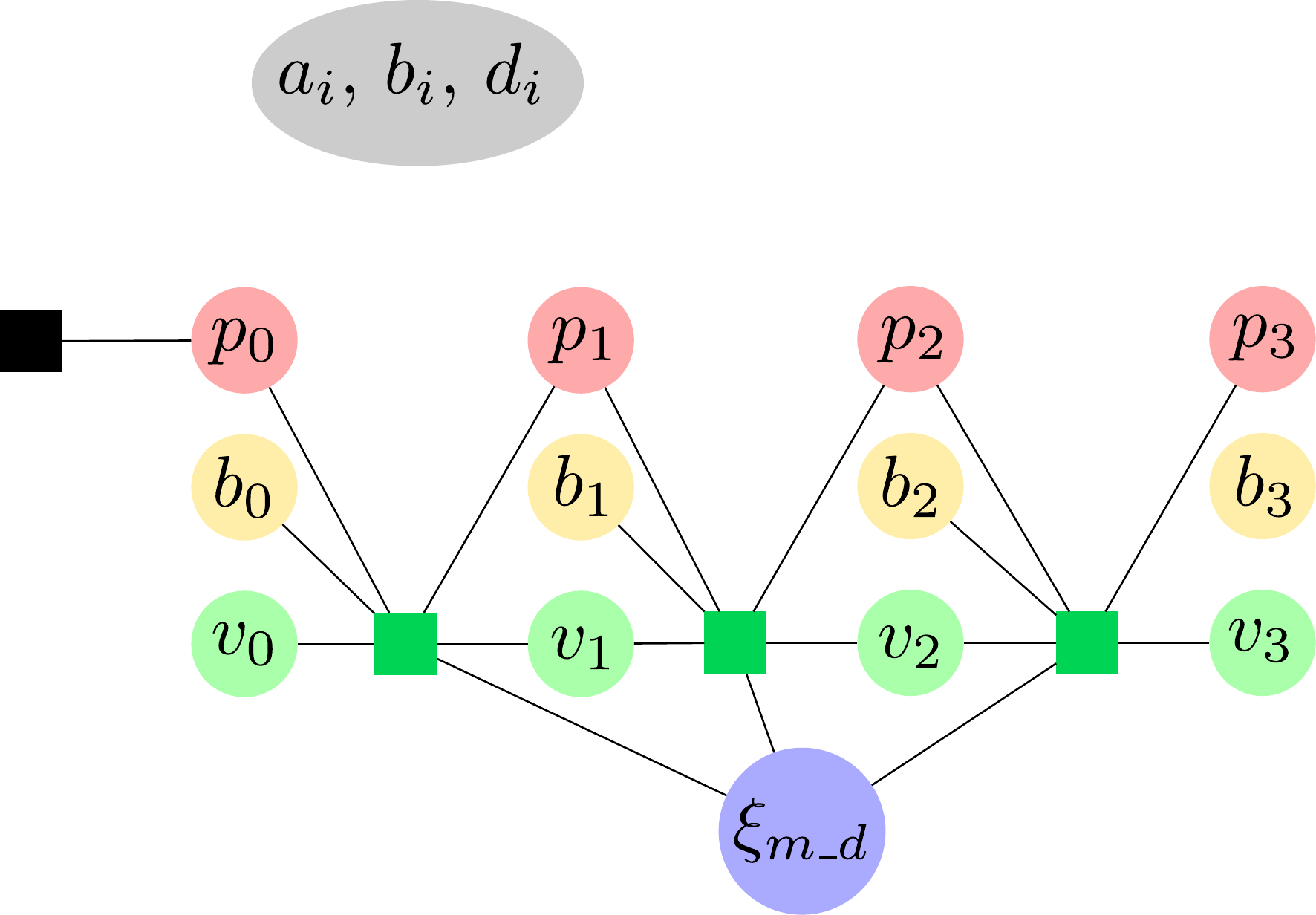}
        \caption{$G_{\text{metric}}$}
        \label{fig:bagraphimu}
    \end{subfigure}~ 
    \begin{subfigure}[t]{0.47\linewidth}
        \includegraphics[width=\linewidth]{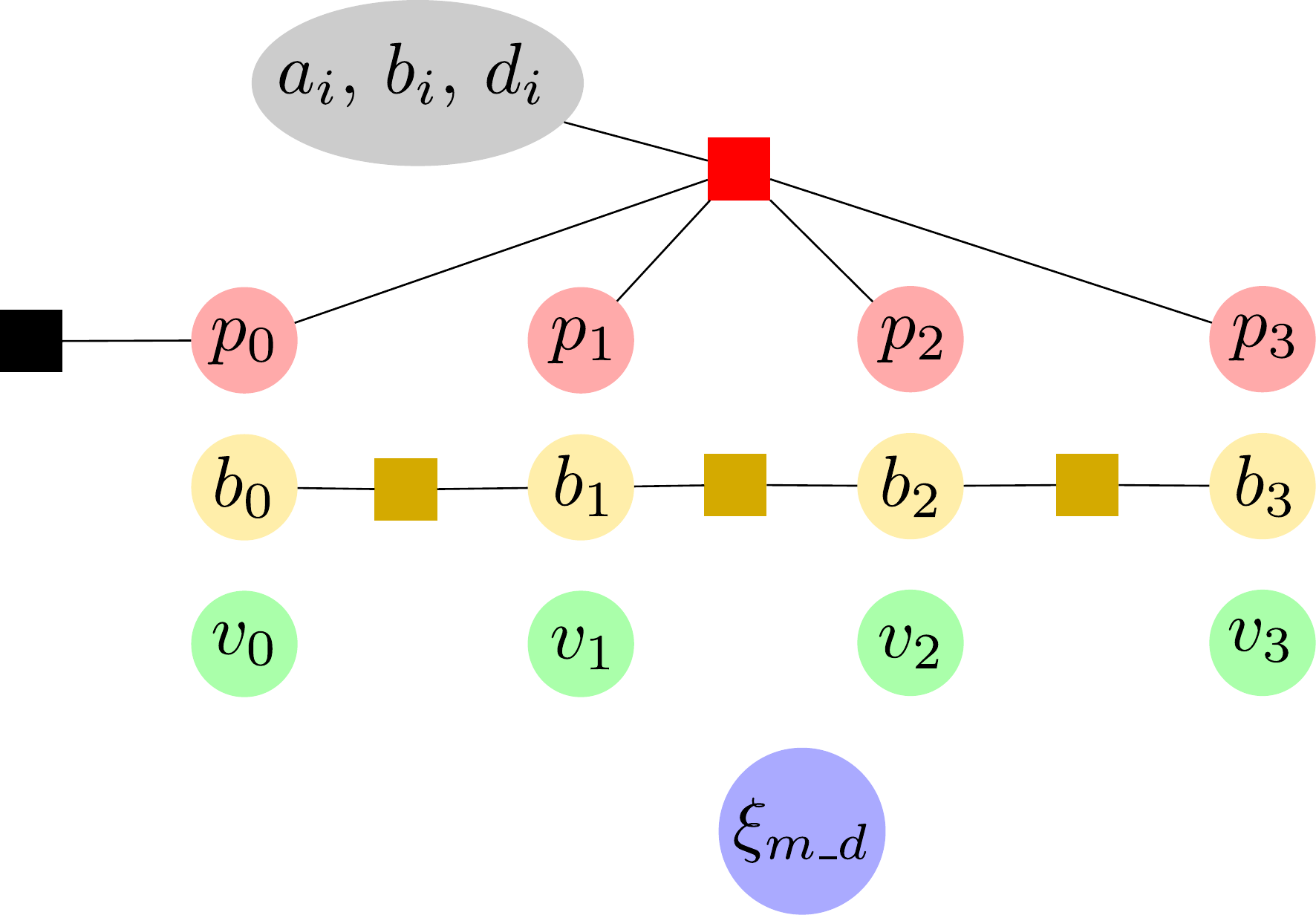}
        \caption{$G_{\text{visual}}$}
        \label{fig:bagraphvisual}
    \end{subfigure}
    \caption{Partitioning of the factor graph from Fig. \ref{fig:bagraph} into $G_{\text{metric}}$ and $G_{\text{visual}}$.  $G_{\text{metric}}$ contains all IMU-factors while $G_{\text{visual}}$ contains the factors that do not depend on $\bm{\xi}_{m\_d}$. Note that both of them do not contain any marginalization factors.}
	\label{fig:bagraphpartition}
\end{figure}

\begin{figure*}[htb]
    \centering
    \includegraphics[width=\linewidth]{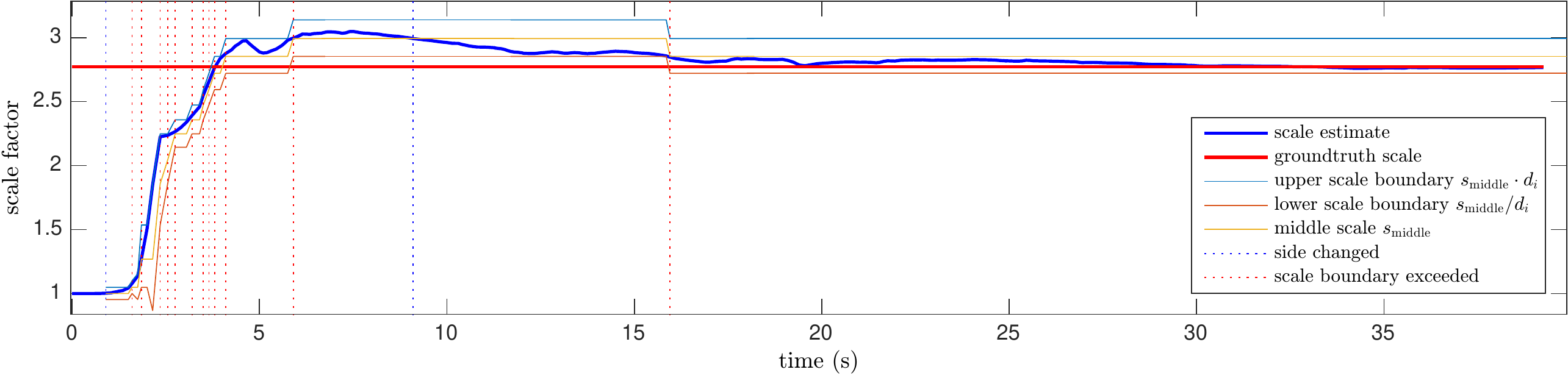}
    \caption{The scale estimation running on the V1\_03\_difficult sequence from the EuRoC dataset. We show the current scale estimate (bold blue), the groundtruth scale (bold red) and the current scale interval (light lines). 
    The vertical dotted lines denote when the side changes (blue) and when the boundary of the scale interval is exceeded (red). In practice this means that $M_{\text{curr}}$ contains the inertial factors since the last blue or red dotted line that is before the last red dotted line. For example at 16s it contains all inertial data since the blue line at 9 seconds.} 
    \label{fig:halfplot} 
\end{figure*}

The marginalization procedure described in subsection \ref{sec:marginalization} has two purposes: reduce the computation complexity of the optimization by removing old states and maintain the information about the previous states of the system. This procedure fixes the linearization points of the states connected to the old states, so they should already have a good estimate. In our scenario this is the case for all variables except of scale. 

The main idea of "Dynamic marginalization" is to maintain several marginalization priors at the same time and reset the one we currently use when the scale estimate moves too far from the linearization point in the marginalization prior.

In our implementation we use three marginalization priors: $M_{\text{visual}}$, $M_{\text{curr}}$ and $M_{\text{half}}$.  $M_{\text{visual}}$ contains only scale independent information from previous states of the vision and cannot be used to infer the global scale. $M_{\text{curr}}$ contains all information since the time we set the linearization point for the scale and $M_{\text{half}}$ contains only the recent states that have a scale close to the current estimate.

When the scale estimate deviates too much from the linearization point of $M_{\text{curr}}$, the value of $M_{\text{curr}}$ is set to $M_{\text{half}}$ and $M_{\text{half}}$ is set to $M_{\text{visual}}$ with corresponding changes in the linearization points. This ensures that the optimization always has some information about the previous states with consistent scale estimates. In the remaining part of the section we provide the details of our implementation.

We define $G_{\text{metric}}$ to contain only the visual-inertial factors (which depend on $\bm{\xi}_{m\_d}$) and $G_{\text{visual}}$ to contain all other factors, except the marginalization priors. Then
\begin{equation}
G_{\text{full}} = G_{\text{metric}} \cup G_{\text{visual}}
\end{equation}Fig. \ref{fig:bagraphpartition} depicts the partitioning of the factor graph.

We define three different marginalization factors $M_{\text{curr}}$, $M_{\text{visual}}$ and $M_{\text{half}}$.
For the optimization we always compute updates using the graph
\begin{equation}
    G_{ba} = G_{\text{metric}} \cup G_{\text{visual}} \cup M_{\text{curr}}
\end{equation}
When keyframe $i$ is marginalized we update $M_{\text{visual}}$ with the factor arising from marginalizing frame $i$ in $G_{\text{visual}} \cup M_{\text{visual}}$. This means that $M_{\text{visual}}$ contains all marginalized visual factors and no marginalized inertial factors making it independent of the scale.

For each marginalized keyframe $i$ we define
\begin{equation}
    s_{i} := \text{scale estimate at the time, $i$ was marginalized}
\end{equation}

We define $i \in M$ if and only if $M$ contains an \emph{inertial} factor that was marginalized at time $i$. Using this we enforce the following constraints for inertial factors. 
\begin{equation} \label{eq:constraintcurr}
    \forall i \in M_{\text{curr}}: s_i \in \left[ s_{\text{middle}} / d_i, s_{\text{middle}} \cdot d_i \right]
\end{equation}
\begin{equation} \label{eq:constrainthalf}
    \forall i \in M_{\text{half}}: s_i \in \begin{cases}
    \left[ s_{\text{middle}}, s_{\text{middle}} \cdot d_i \right], &\text{if } s_{\text{curr}} > s_{\text{middle}} \\
    \left[ s_{\text{middle}} / d_i, s_{\text{middle}} \right], &\text{otherwise}
    \end{cases}
\end{equation}

where $s_{\text{middle}}$ is the current middle of the allowed scale interval (initialized with $s_0$), $d_i$ is the size of the scale interval at time $i$, and $s_{\text{curr}}$ is the current scale estimate.

We update $M_{\text{curr}}$ by marginalizing frame $i$ in $G_{\text{ba}}$ and we update $M_{\text{half}}$ by marginalizing $i$ in $G_{\text{metric}} \cup G_{\text{visual}} \cup M_{\text{half}}$
 
 In order to preserve the constraints in Equations (\ref{eq:constraintcurr}) and (\ref{eq:constrainthalf}) we apply Algorithm \ref{alg:constrainmarg} everytime a marginalization happens.
By following these steps on the one hand we make sure that the constraints are satisfied which ensures that the scale difference in the currently used marginalization factor stays smaller than $d_i^2$. On the other hand the factor always contains some inertial factors so that the scale estimation works at all times. Note also that $M_{\text{curr}}$ and $M_{\text{half}}$ have separate First Estimate Jacobians that are employed when the respective marginalization factor is used. Fig. \ref{fig:halfplot} shows how the system works in practice.

\begin{algorithm}
\caption{Constrain Marginalization} \label{alg:constrainmarg}
    \begin{algorithmic}
    \State $\text{upper} \gets s_{\text{curr}} > s_{\text{middle}}$
    \If{$\text{upper} \neq \text{lastUpper}$} \Comment{Side changes.}
        \State $M_{\text{half}} \gets M_{\text{visual}}$
    \EndIf
    \If{$s_{\text{curr}} > s_{\text{middle}} \cdot d_i$} \Comment{Upper boundary exceeded.}
        \State $M_{\text{curr}} \gets M_{\text{half}}$
        \State $M_{\text{half}} \gets M_{\text{visual}}$
        \State $s_{\text{middle}} \gets s_{\text{middle}} \cdot d_i$
    \EndIf
    \If{$s_{\text{curr}} < s_{\text{middle}} / d_i$} \Comment{Lower boundary exceeded.}
        \State $M_{\text{curr}} \gets M_{\text{half}}$
        \State $M_{\text{half}} \gets M_{\text{visual}}$
        \State $s_{\text{middle}} \gets s_{\text{middle}} / d_i$
    \EndIf
    \State $\text{lastUpper} \gets \text{upper}$
    \end{algorithmic}
\end{algorithm} 

An important part of this strategy is the choice of $d_i$. It should be small, in order to keep the system consistent, but not too small so that $M_{\text{curr}}$ always contains enough inertial factors. Therefore we chose to dynamically adjust the parameter as follows. At all time steps $i$ we calculate
\begin{equation}
    d_{i} =\min{ \{ d_{\text{min}}^j ~|~ j \in \mathbb{N} \setminus \{ 0 \} , \frac{s_i}{s_{i-1}} < d_i \} }
\end{equation}
This ensures that it cannot happen that the $M_{\text{half}}$ gets reset to $M_{\text{visual}}$ at the same time that $M_{\text{curr}}$ is exchanged with $M_{\text{half}}$. Therefore it prevents situations where $M_{\text{curr}}$ contains no inertial factors at all, making the scale estimation more reliable. In our experiments we chose $d_{\text{min}} = \sqrt{1.1}$.

\begin{figure}[tb]
    \centering
    \begin{subfigure}[t]{0.49\linewidth}
        \includegraphics[width=\linewidth]{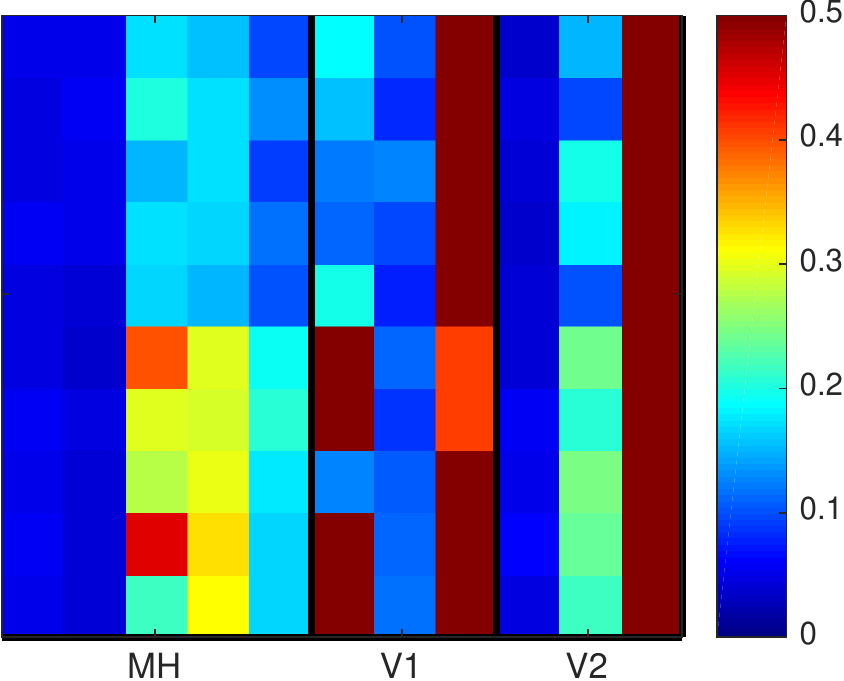}
        \caption{DSO \cite{engel2016dso} realtime gt-scaled}
        \label{fig:dsort}
    \end{subfigure}~
    \begin{subfigure}[t]{0.49\linewidth}
        \includegraphics[width=\linewidth]{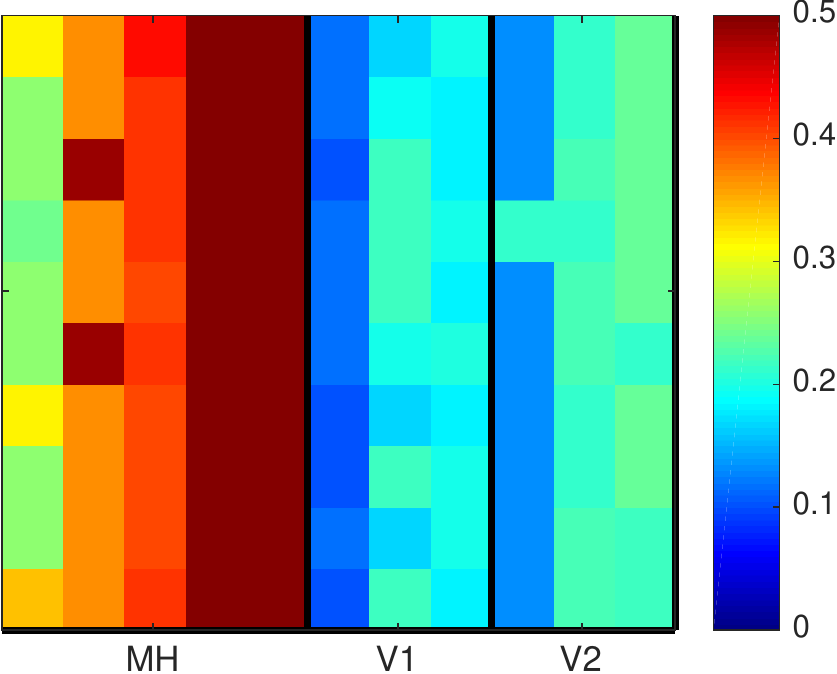}
        \caption{ROVIO realtime \cite{Bloesch} }
        \label{fig:rovio}
    \end{subfigure} \\
    \begin{subfigure}[t]{0.49\linewidth}
        \includegraphics[width=\linewidth]{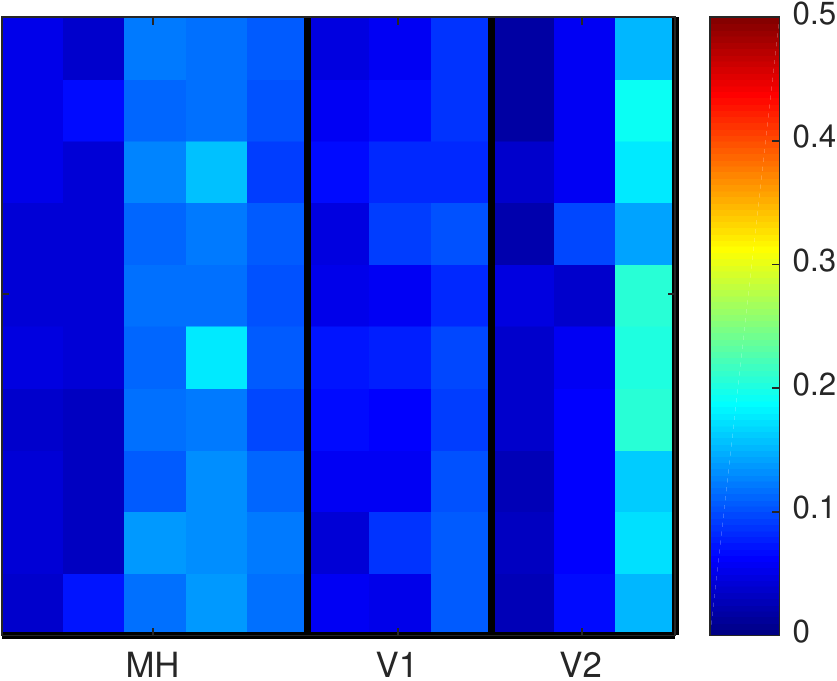}
        \caption{our method realtime gt-scaled}
        \label{fig:squareourgtscaled}
    \end{subfigure}~
    \begin{subfigure}[t]{0.49\linewidth}
        \includegraphics[width=\linewidth]{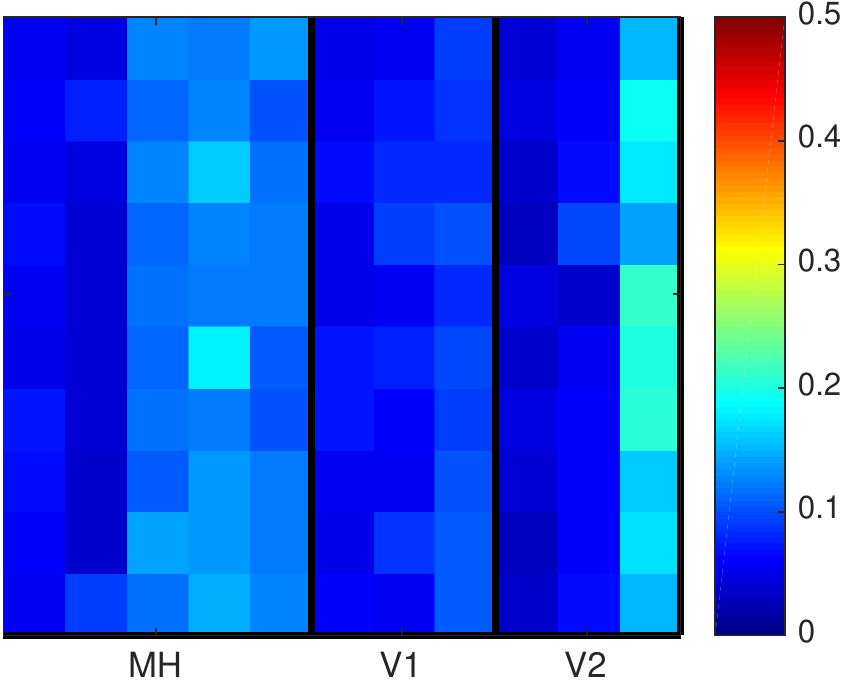}
        \caption{our method realtime}
        \label{fig:squareourunscaled}
    \end{subfigure}
    \caption{rmse for different methods run 10 times (lines) on each sequence (columns) of the EuRoC dataset. 
    }
	\label{fig:squareplots}
\end{figure} 

\begin{figure}[tb]
    \centering
    \includegraphics[width=\linewidth]{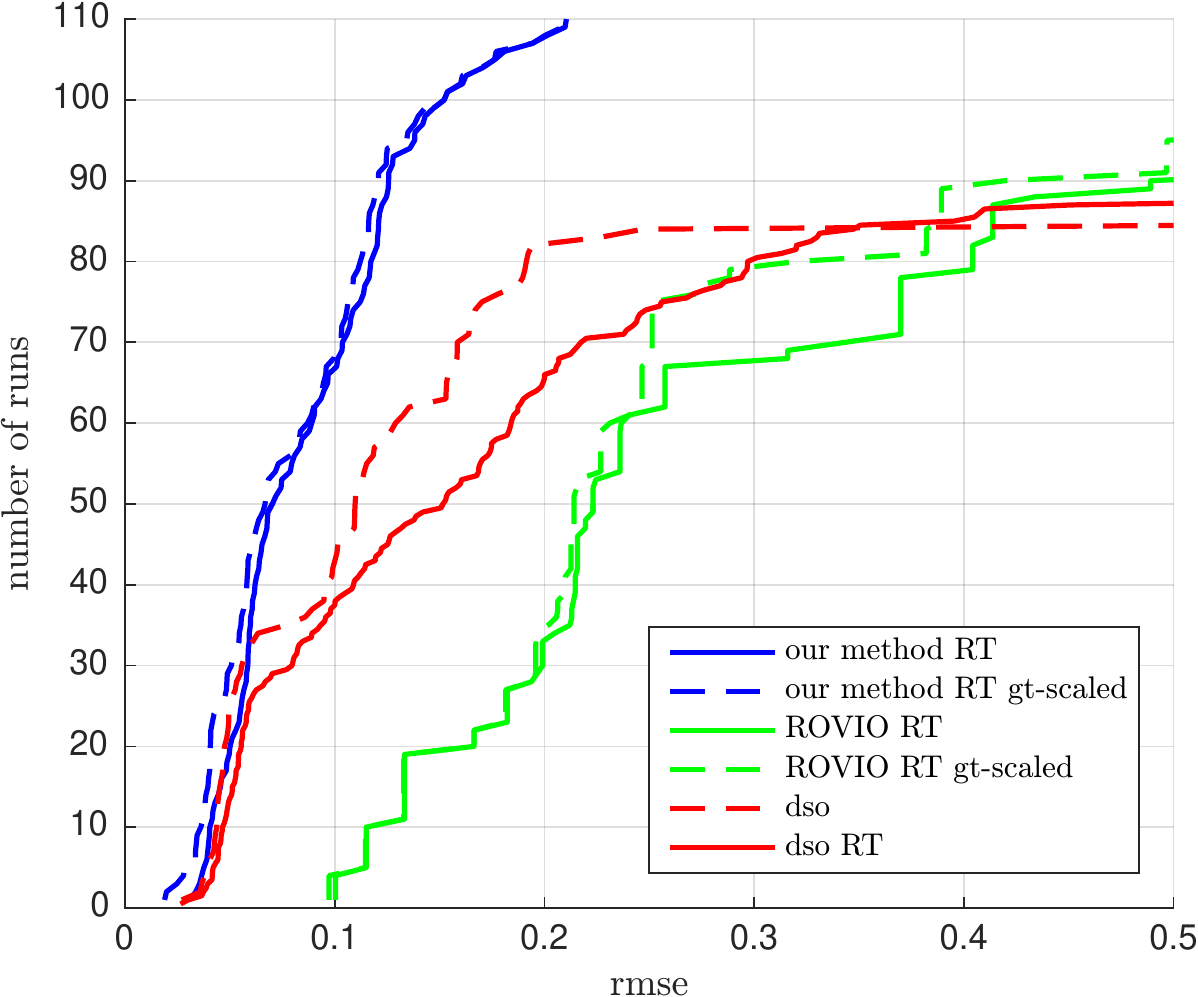} 
    \caption{Cumulative error plot on the EuRoC-dataset (RT means realtime). 
    This experiment demonstrates that the additional IMU not only provides a reliable scale estimate, but that it also significantly increases accuracy and robustness.}
    \label{fig:lineplots1}
\end{figure}

\subsection{Coarse Visual-Inertial Tracking}

The coarse tracking is responsible for computing a fast pose estimate for each frame that also serves as an initialization for the joint optimization detailed in \ref{sec:jointoptimization}. We perform conventional direct image alignment between the current frame and the latest keyframe, while keeping the geometry and the scale fixed. 
Inertial residuals using the previously described IMU preintegration scheme are placed between subsequent frames.
Everytime the joint optimization is finished for a new frame, the coarse tracking is reinitialized with the new estimates for scale, gravity direction, bias, and velocity as well as the new keyframe as a reference for the visual factors. Similar to the joint optimization we perform partial marginalization to keep the update time constrained. After estimating the variables for a new frame we marginalize out all variables except the keyframe pose and the variables of the newest frame. 
In contrast to the joint optimization we do not need to use dynamic marginalization because the scale is not included in the optimization.

\section{Results}

We evaluate our approach on the publicly available EuRoC dataset \cite{Burri2016Euroc}. 
The performance is compared to \cite{engel2016dso}, \cite{Bloesch}, \cite{murartal15orbslam}, \cite{usenko16icra}, \cite{leutenegger2014keyframe} and \cite{Kasyanov2017_VISLAM}. We also provide supplementary material with more evaluation and a video at \href{http://vision.in.tum.de/vi-dso}{vision.in.tum.de/vi-dso}.

\subsection{Robust Quantitative Evaluation}

In order to obtain an accurate evaluation we run our method 10 times for each sequence of the dataset (using the left camera). We directly compare the results to visual-only DSO \cite{engel2016dso} and ROVIO \cite{Bloesch}.
As DSO cannot observe the scale we evaluate using the optimal ground truth scale in some plots (with the description "gt-scaled") to enable a fair comparison. For all other results we scale the trajectory with the final scale estimate (our method) or with $1$ (other methods).
For DSO we use the results published together with their paper. We use the same start and end times for each sequence to run our method and ROVIO. Note that the drone has a high initial velocity in some sequences when using these start times making it especially challenging for our IMU initialization.
Fig. \ref{fig:squareplots} shows the root mean square error (rmse) for every run and Fig. \ref{fig:lineplots1} displays the cumulative error plot. Clearly our method significantly outperforms DSO and ROVIO. Without inertial data DSO is not able to work on all sequences especially on V1\_03\_difficult and V2\_03\_difficult and it is also not able to scale the results correctly. ROVIO on the other hand is very robust but as a filtering-based method it cannot provide sufficient accuracy.

\begin{table*}[ht]
\caption{Accuracy of the estimated trajectory on the EuRoC dataset for several methods. 
Note that ORB-SLAM does a convincing job showing leading performance on some of the sequences. Nevertheless, since our method directly works on the sensor data (colors and IMU measurements), we observe similar precision and a better robustness -- even without loop closuring. Moreover, the proposed method is the only one not to fail on any of the sequences.
}
\label{tab:euroc}
\begin{center}
  \begin{tabular}{c c|c c c c c c c c c c c}
    \toprule
Sequence &  & MH1 & MH2 & MH3 & MH4 & MH5 & V11 & V12 & V13 & V21 & V22 & V23 \\
\midrule
\multirow{3}{*}{\begin{tabular}{c}VI-DSO (our method, RT)\\ (median of 10 runs each)\end{tabular}} & RMSE & \textbf{0.062} & \textbf{0.044} & 0.117 & \textbf{0.132} & 0.121 & 0.059 & 0.067 & \textbf{0.096} & 0.040 & 0.062 & 0.174 \\
 & RMSE gt-scaled & 0.041 & 0.041 & 0.116 & 0.129 & 0.106 & 0.057 & 0.066 & 0.095& 0.031 & 0.060 & 0.173 \\
 & Scale Error (\%) & 1.1 & \textbf{0.5} & \textbf{0.4} & \textbf{0.2} & 0.8 & 1.1 & 1.1 & \textbf{0.8} & 1.2 & \textbf{0.3} & \textbf{0.4} \\
 \midrule
\multirow{3}{*}{\begin{tabular}{c} VI ORB-SLAM \\ (keyframe trajectory) \end{tabular}} & RMSE & 0.075 & 0.084 & \textbf{0.087} & 0.217 & \textbf{0.082} & \textbf{0.027} & \textbf{0.028} & X & \textbf{0.032} & \textbf{0.041} & \textbf{0.074} \\
& RMSE gt-scaled & 0.072 & 0.078 & 0.067 & 0.081 & 0.077 & 0.019 & 0.024 & X & 0.031 & 0.026 & 0.073 \\
& Scale Error (\%) & \textbf{0.5} & 0.8 & 1.5 & 3.4 & \textbf{0.5} & \textbf{0.9} & \textbf{0.8} & X & \textbf{0.2} & 1.4 & 0.7 \\
\midrule
VI odometry \cite{leutenegger2014keyframe}, mono & RMSE & 0.34 & 0.36 & 0.30 & 0.48 & 0.47 & 0.12 & 0.16 & 0.24 & 0.12 & 0.22 & X \\ 
VI odometry \cite{leutenegger2014keyframe}, stereo & RMSE & 0.23 & 0.15 & 0.23 & 0.32 & 0.36 & 0.04 & 0.08 & 0.13 & 0.10 & 0.17 & X \\
\midrule
VI SLAM \cite{Kasyanov2017_VISLAM}, mono & RMSE & 0.25 & 0.18 & 0.21 & 0.30 & 0.35 & 0.11 & 0.13 & 0.20 & 0.12 & 0.20 & X \\
VI SLAM \cite{Kasyanov2017_VISLAM}, stereo & RMSE & 0.11 & 0.09 & 0.19 & 0.27 & 0.23 & 0.04 & 0.05 & 0.11 & 0.10 & 0.18 & X \\
\bottomrule
  \end{tabular}
\end{center} 
\end{table*}

Table \ref{tab:euroc} shows a comparison to several other methods. For our results we have displayed the median error for each sequence from the 10 runs plotted in Fig. \ref{fig:squareourgtscaled}. This makes the results very meaningful. For the other methods unfortunately only one result was reported so we have to assume that they are representative as well.
The results for \cite{leutenegger2014keyframe} and \cite{Kasyanov2017_VISLAM} were taken from \cite{Kasyanov2017_VISLAM}. 
The results for \cite{murartal15orbslam} (as reported in their paper) differ slightly from the other methods as they show the error of the keyframe trajectory instead of the full trajectory. This is a slight advantage as keyframes are bundle-adjusted in their method which does not happen for the other frames.

 In comparison to VI ORB-SLAM our method outperforms it in terms of rmse on several sequences. As ORB-SLAM is a SLAM system while ours is a pure odometry method this is a remarkable achievement especially considering the differences in the evaluation. Note that the Vicon room sequences (V*) are executed in a small room and contain a lot of loopy motions where the loop closures done by a SLAM system significantly improve the performance.
 Also our method is more robust as ORB-SLAM fails to track one sequence. Even considering only sequences where ORB-SLAM works our approach has a lower maximum rmse.

Compared to \cite{leutenegger2014keyframe} and \cite{Kasyanov2017_VISLAM} our method obviously outperforms them. It is better than the monocular versions on every single sequence and it beats even the stereo and SLAM-versions on 9 out of 11 sequences.

In summary our method is the only one which is able to track all the sequences successfully except ROVIO.

\begin{figure*}[htb]
    \captionsetup[subfigure]{aboveskip=2pt}
    \centering
    \begin{subfigure}[t]{0.4\linewidth}
        \includegraphics[width=\linewidth]{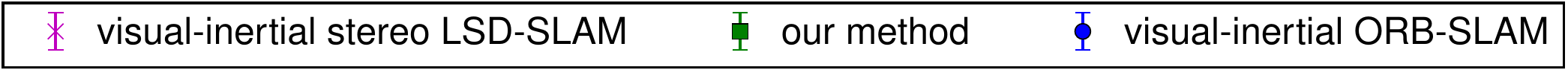}
    \end{subfigure} \\
    \vspace{0.5mm}
    \begin{subfigure}[b]{0.3\linewidth} 
    \centering
        \includegraphics[width=\linewidth]{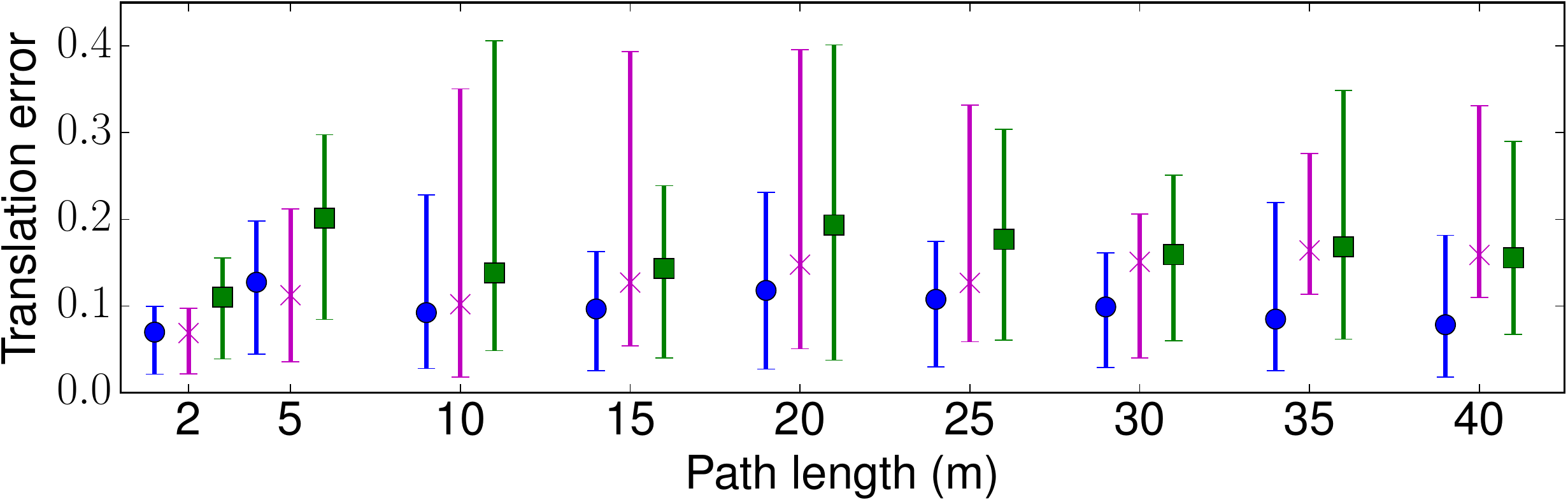}
        \caption{Translation error V1\_01\_easy}
        \label{fig:lsdplot2}
    \end{subfigure}~
    \begin{subfigure}[b]{0.3\linewidth}
    \centering
        \includegraphics[width=\linewidth]{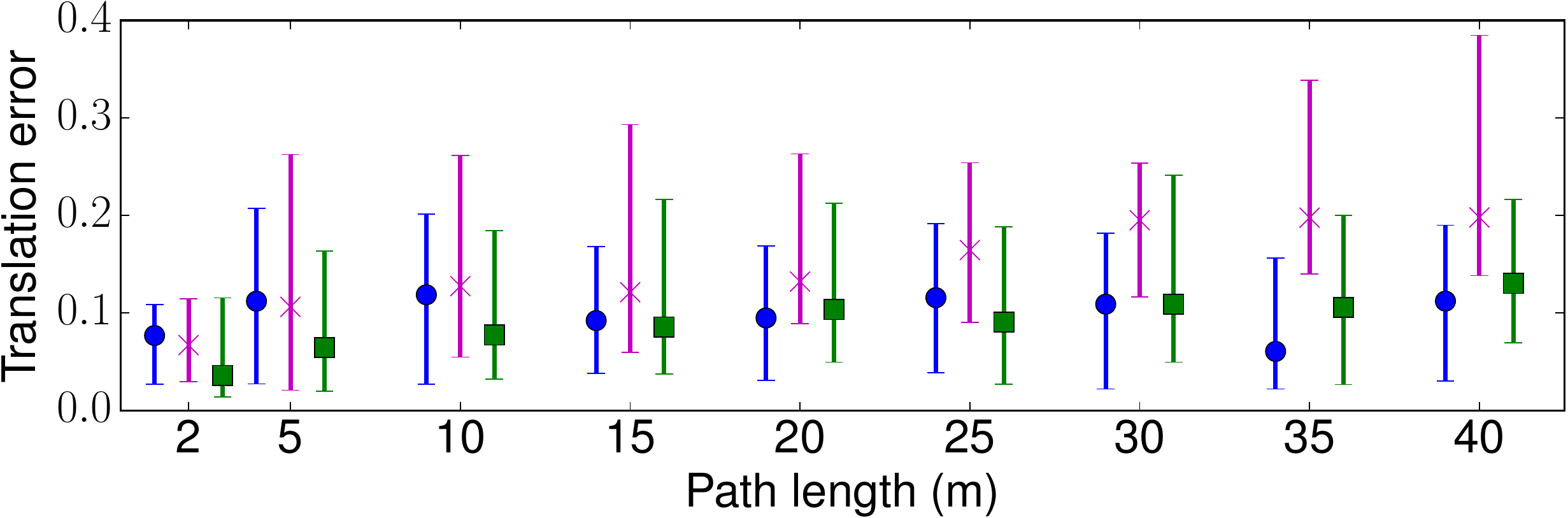}
        \caption{Translation error V1\_02\_medium}
        \label{fig:lsdplot4}
    \end{subfigure}~
    \begin{subfigure}[b]{0.3\linewidth}
    \centering
        \includegraphics[width=\linewidth]{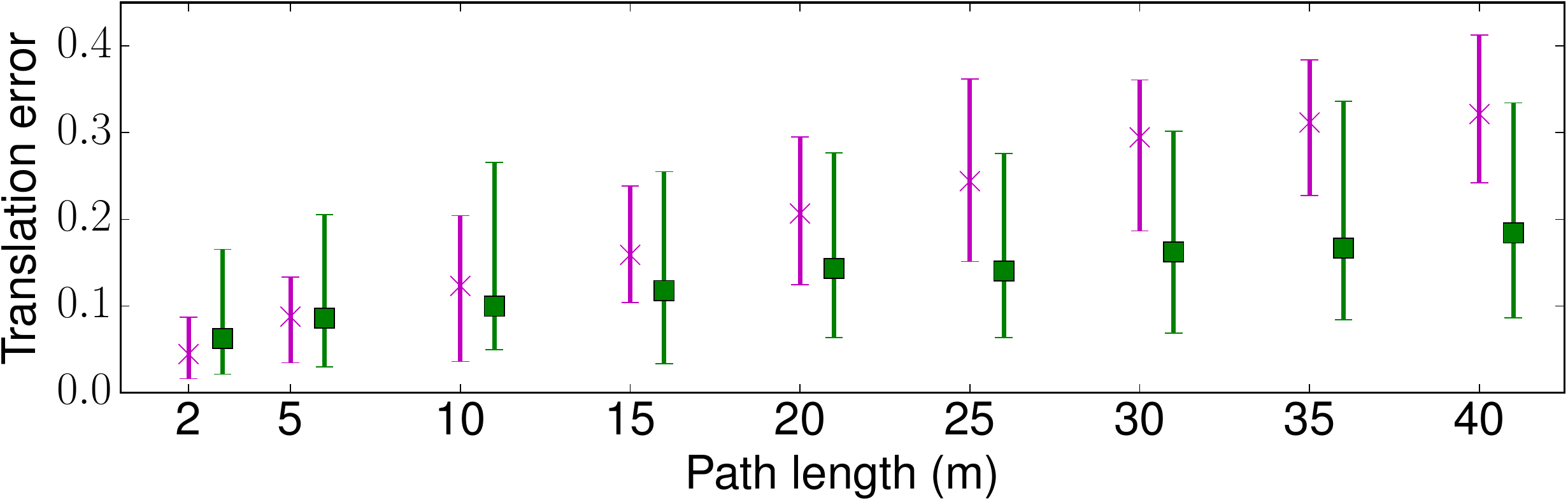}
        \caption{Translation error V1\_03\_difficult}
        \label{fig:lsdplot6}
    \end{subfigure}
    \begin{subfigure}[b]{0.3\linewidth}
    \centering
        \includegraphics[width=\linewidth]{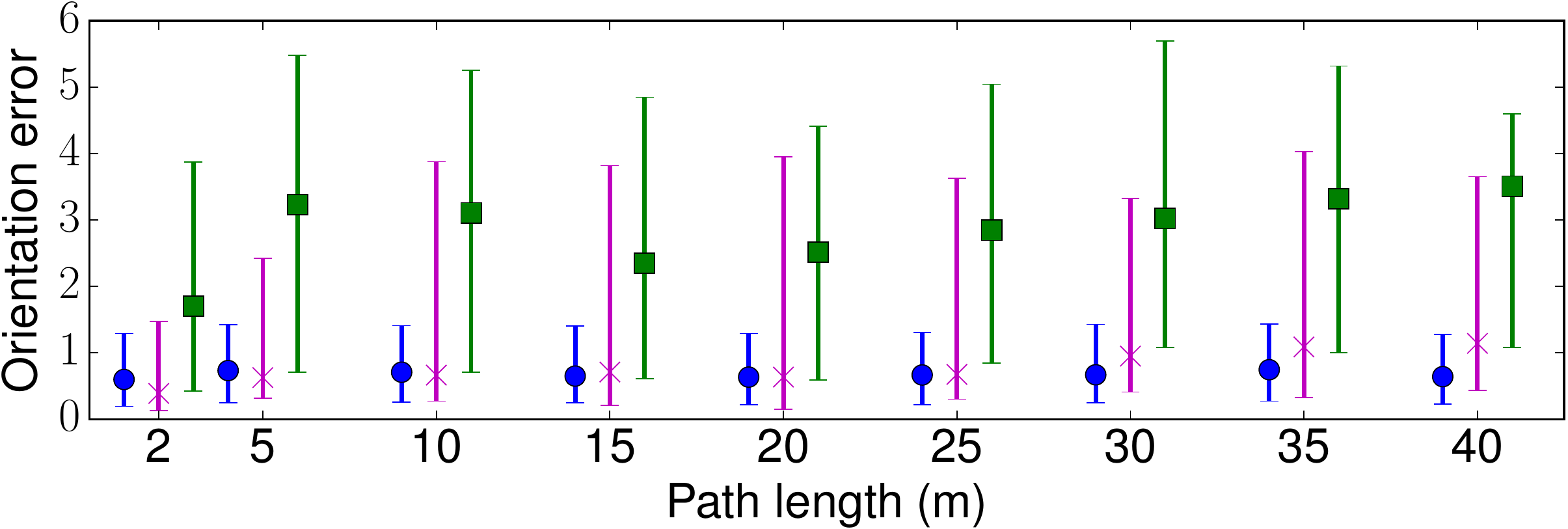}
        \caption{Orientation error V1\_01\_easy}
        \label{fig:lsdplot1}
    \end{subfigure}~
    \begin{subfigure}[b]{0.3\linewidth}
    \centering
        \includegraphics[width=\linewidth]{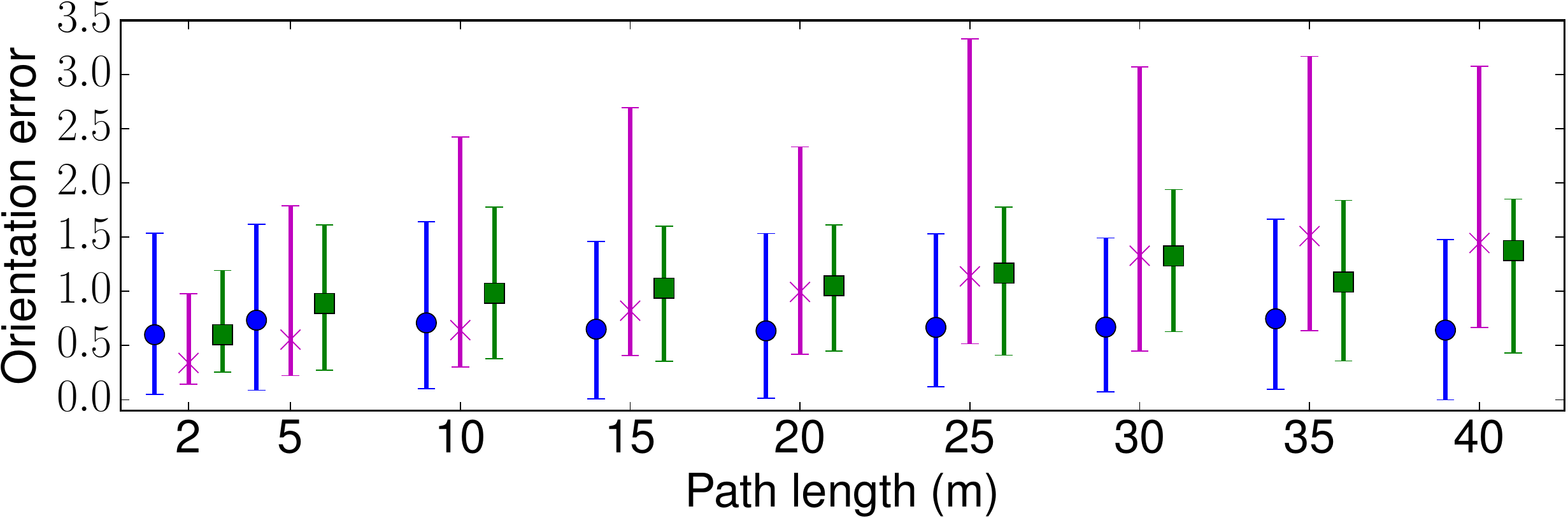}
        \caption{Orientation error V1\_02\_medium}
        \label{fig:lsdplot3}
    \end{subfigure}~
    \begin{subfigure}[b]{0.3\linewidth}
    \centering
        \includegraphics[width=\linewidth]{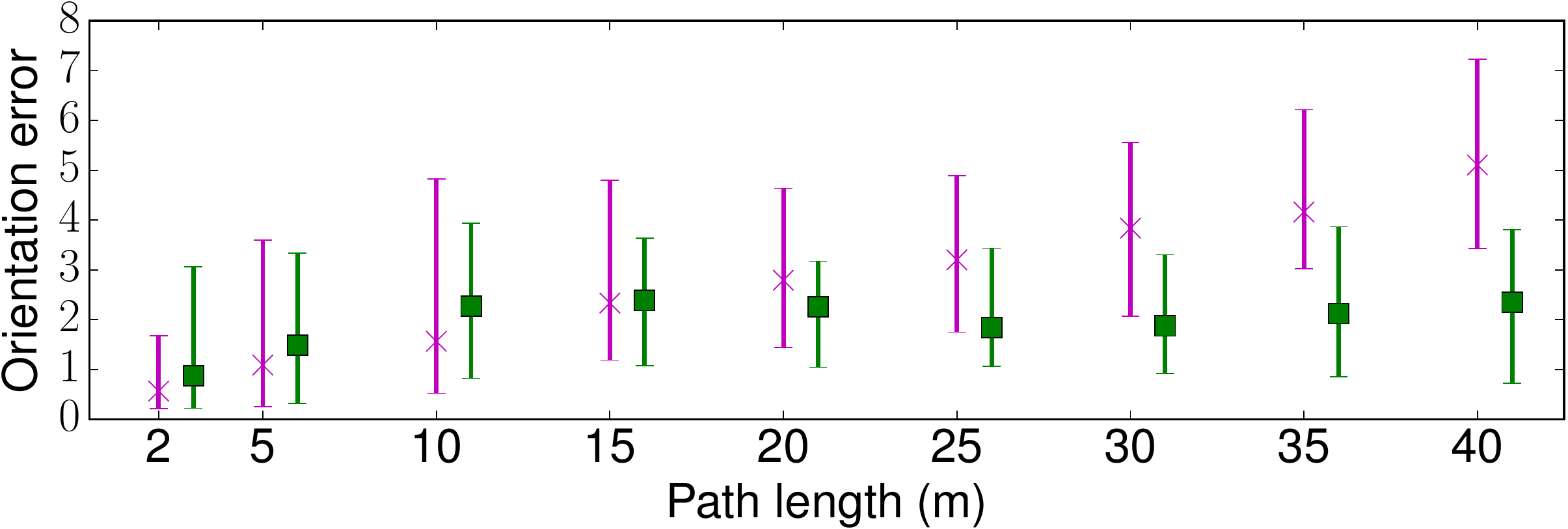}
        \caption{Orientation error V1\_03\_difficult}
        \label{fig:lsdplot5}
    \end{subfigure}
    \caption{Relative Pose Error evaluated on three sequences of the EuRoC-dataset for visual-inertial ORB-SLAM \cite{murartal15orbslam}, visual-inertial stereo LSD-SLAM \cite{usenko16icra} and our method. 
    Although the proposed VI-DSO does not use loop closuring (like \cite{murartal15orbslam}) or stereo (like \cite{usenko16icra}), VI-DSO is quite competitive in terms of accuracy and robustness.  Note that \cite{murartal15orbslam} with loop closures is slightly more accurate on average, yet it entirely failed on V1\_03\_difficult.
    }
	\label{fig:lsdplots}
\end{figure*}

\begin{figure}[tb]
    \centering
    \includegraphics[height=3cm]{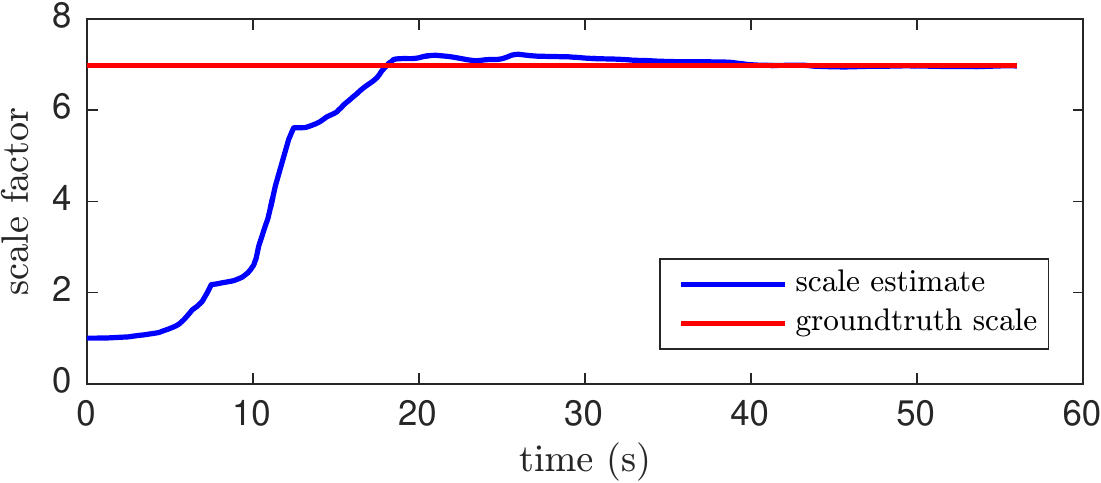}
    \caption{Scale estimate for MH\_04\_difficult (median result of 10 runs in terms of tracking accuracy). Note how the estimated scale converges to the correct value despite being initialized far from the optimum.}
	\label{fig:scaleplot}
\end{figure}

We also compare the Relative Pose Error to \cite{murartal15orbslam} and \cite{usenko16icra} on the V1\_0*-sequences of EuRoC (Fig. \ref{fig:lsdplots}). While our method cannot beat the SLAM system and the stereo method on the easy sequence we outperform \cite{usenko16icra} and are as good as \cite{murartal15orbslam} on the medium sequence. On the hard sequence we outperform both of the contenders even though we neither use stereo nor loop-closures.

\subsection{Evaluation of the Initialization}

There are only few methods we can compare our initialization to. Some approaches like \cite{Martinelli2014} have not been tested on real data. While \cite{Kaiser14imu_init} provides results on real data, the dataset used was featuring a downward-looking camera and an environment with a lot of features which is not comparable to the EuRoC-dataset in terms of difficulty. Also they do not address the problem of late observability which suggests that a proper motion is performed in the beginning of their dataset.
As a filtering-based method ROVIO does not need a specific initialization procedure but it also cannot compete in terms of accuracy making it less relevant for this discussion.
Visual-inertial LSD-SLAM uses stereo and therefore does not face the main problem of scale estimation. 
Therefore we compare our initialization procedure to visual-inertial ORB-SLAM \cite{murartal15orbslam} as both of the methods work on the challenging EuRoC-dataset and have to estimate the scale, gravity direction, bias, and velocity.

In comparison to \cite{murartal15orbslam} our estimated scale is better overall (Table \ref{tab:euroc}). On most sequences our method provides a better scale, and our average scale error ($0.7\%$ compared to $1.0\%$) as well as our maximum scale error ($1.2\%$ compared to $3.4\%$) is lower. In addition our method is more robust as the initialization procedure of \cite{murartal15orbslam} fails on V1\_03\_difficult.

Apart from the numbers we argue that our approach is superior in terms of the general structure. While \cite{murartal15orbslam} have to wait for 15 seconds until the initialization is performed, our method provides an approximate scale and gravity direction almost instantly, that gets enhanced over time. Whereas in \cite{murartal15orbslam} the pose estimation has to work for 15 seconds without any IMU data, in our method the inertial data is used to improve the pose estimation from the beginning. 
This is probably one of the reasons why our method is able to process V1\_03\_difficult. 
Finally our method is better suited for robotics applications. For example an autonomous drone is not able to fly without gravity direction and scale for 15 seconds and hope that afterwards the scale was observable. In contrast our method offers both of them right from the start. The continuous rescaling is also not a big problem as an application could use the unscaled measurements for building a consistent map and for providing flight goals, whereas the scaled measurements can be used for the controller. 
Fig. \ref{fig:scaleplot} shows the scale estimation for MH\_04.

Overall we argue that our initialization procedure exceeds the state of the art and think that the concept of initialization with a very rough scale estimate and jointly estimating it during pose estimation will be a useful concept in the future.

\section{Conclusion}
We have presented a novel formulation of direct sparse visual-inertial odometry. 
We explicitely include scale and gravity direction in our model in order to deal with cases where the scale is not immediately observable. As the initial scale can be very far from the optimum we have proposed a novel technique called dynamic marginalization where we maintain multiple marginalization priors and constrain the maximum scale difference. 
Extensive quantitative evaluation demonstrates that the proposed visual-inertial odometry method outperforms the state of the art, both the complete system as well as the IMU initialization procedure. In particular, experiments confirm that the inertial information not only provides a reliable scale estimate, but it also drastically increases precision and robustness.

\section*{Acknowledgements}
We thank Jakob Engel for releasing the code of DSO and for his helpful comments on First Estimates Jacobians, and the authors of \cite{murartal15orbslam} for providing their numbers for the comparison in Fig. \ref{fig:lsdplots}. 

\def\url#1{}
\bibliographystyle{IEEEtranS}
\bibliography{references}

\end{document}